\documentclass{article}
\pdfoutput=1
\usepackage{nips14submit_e,times}
\usepackage{hyperref}
\usepackage{url}
\usepackage{graphicx}
\usepackage{amsmath}
\usepackage{amssymb}
\usepackage{amsthm}
\usepackage{subfig}

\DeclareMathOperator*{\fop}{FoP}
\DeclareMathOperator*{\data}{data}

\title{Multiscale Fields of Patterns}

\author{
Pedro F. Felzenszwalb \\
Brown University \\
Providence, RI 02906 \\
\texttt{pff@brown.edu} \\
\And
John G. Oberlin \\
Brown University \\
Providence, RI 02906 \\
\texttt{john\_oberlin@brown.edu}}

\nipsfinalcopy 

\begin{document}

\maketitle

\begin{abstract}
We describe a framework for defining high-order image models
that can be used in a variety of applications.  The approach involves
modeling local patterns in a multiscale
representation of an image.
Local properties of a coarsened image reflect non-local properties of
the original image.  In the case of binary images local properties are
defined by the binary patterns observed over small neighborhoods
around each pixel.  With the multiscale representation we capture the
frequency of patterns observed at different scales of resolution.  This
framework leads to expressive priors that depend on a relatively small
number of parameters.  For inference and learning we use an MCMC
method for block sampling with very large blocks.  We evaluate the
approach with two example applications.  One involves contour
detection.  The other involves binary segmentation.
\end{abstract}

\section{Introduction}

Markov random fields are widely used as priors for solving a variety
of vision problems such as image restoration and stereo
\cite{BVZ01,FH06}.  Most of the work in the area has concentrated on
low-order models involving pairs of neighboring pixels.  However, it
is clear that realistic image priors need to capture higher-order
properties of images.

In this paper we describe a general framework
for defining high-order image models that can be used in a variety of
applications.  The approach involves modeling local properties in a
multiscale representation of an image.  This leads to a natural
low-dimensional representation of a high-order model.
We concentrate on the problem of estimating binary images.  In this
case local image properties can be captured by the binary patterns in
small neighborhoods around each pixel.

\begin{figure}
\centering
\begin{tabular}{cccc}
\shortstack{
\fbox{\includegraphics[width=0.325in]{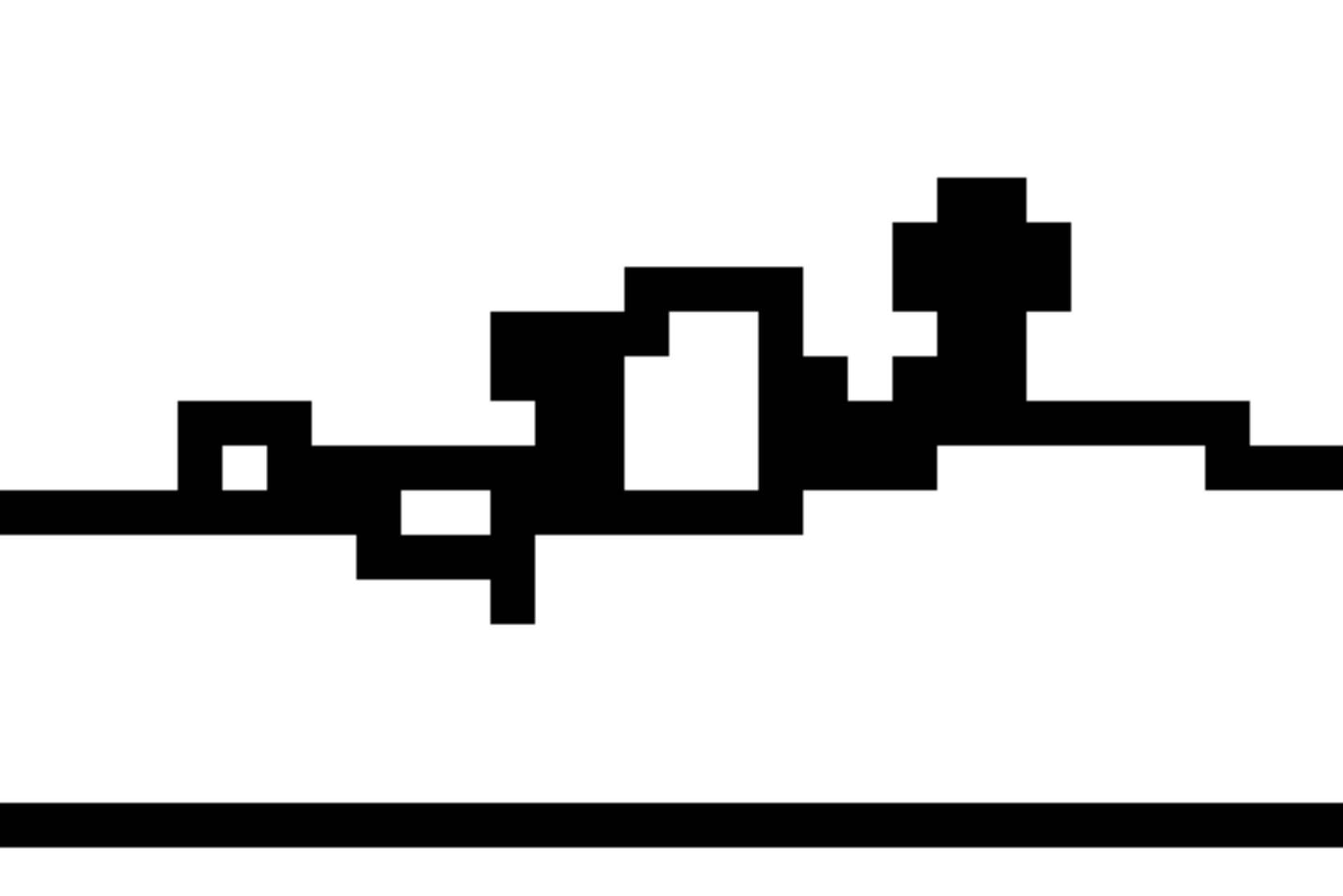}} \\
\fbox{\includegraphics[width=0.65in]{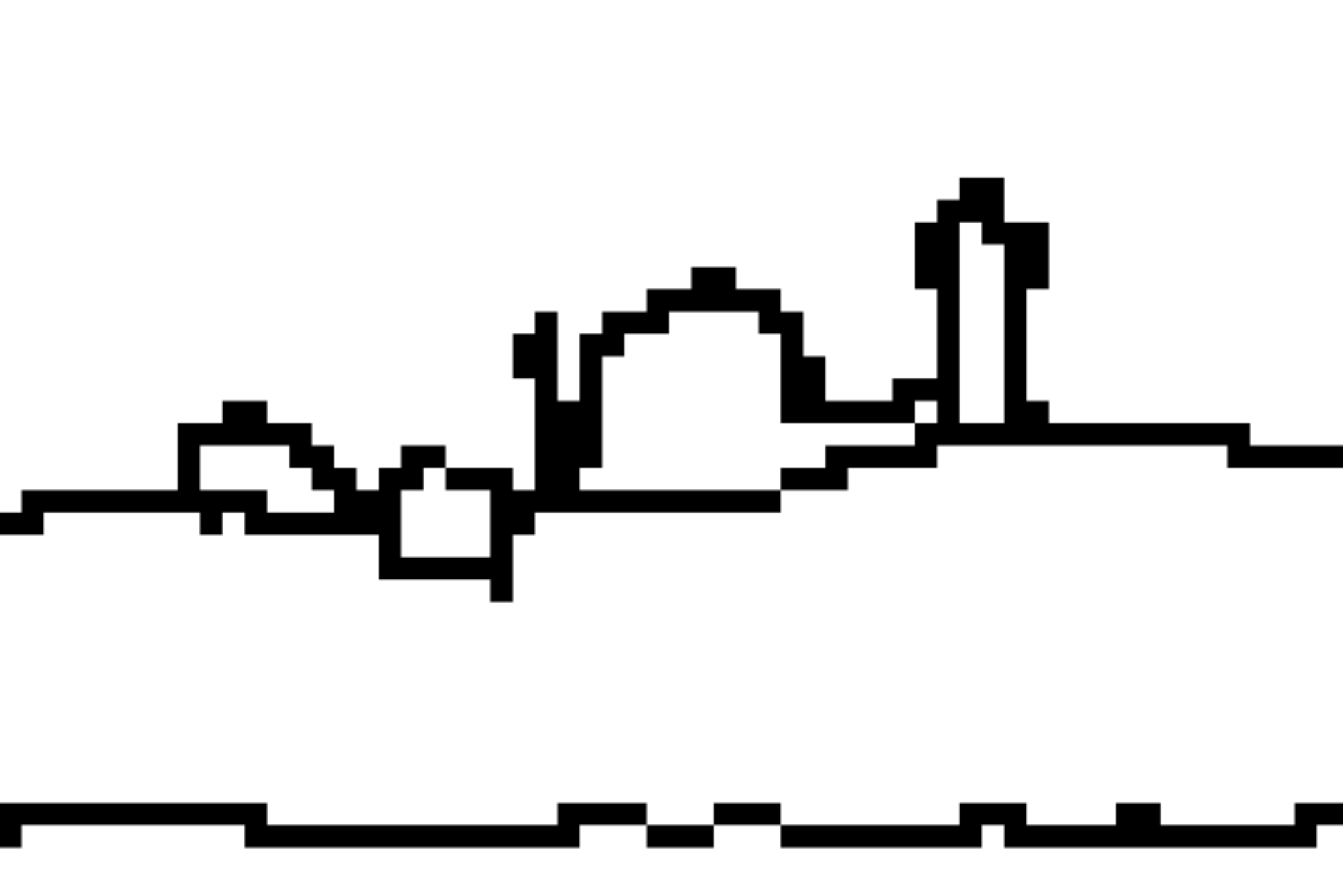}} \\
\fbox{\includegraphics[width=1.3in]{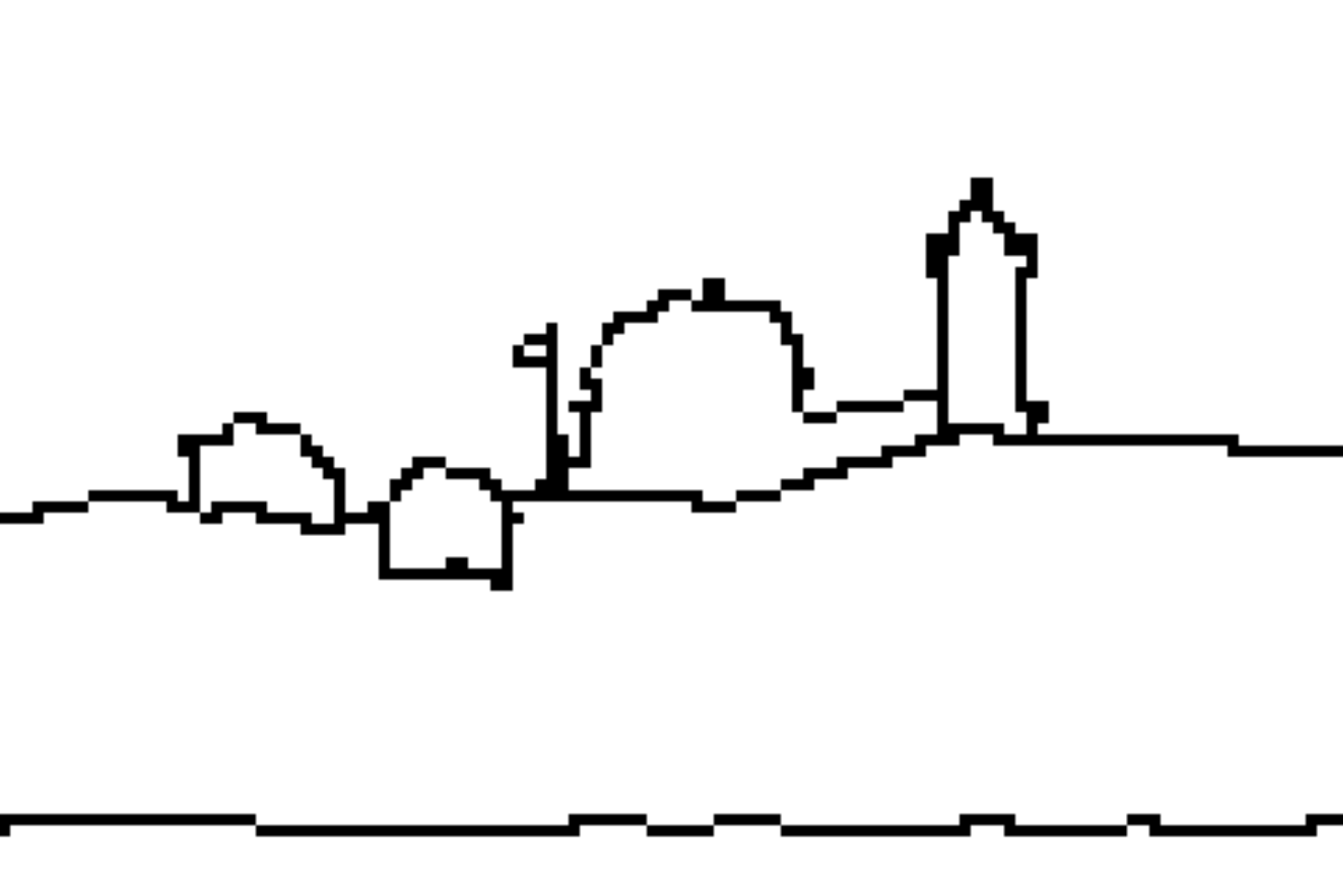}}} &
\shortstack{
\includegraphics[width=0.3in]{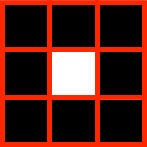} \\
\fbox{\includegraphics[width=1.3in]{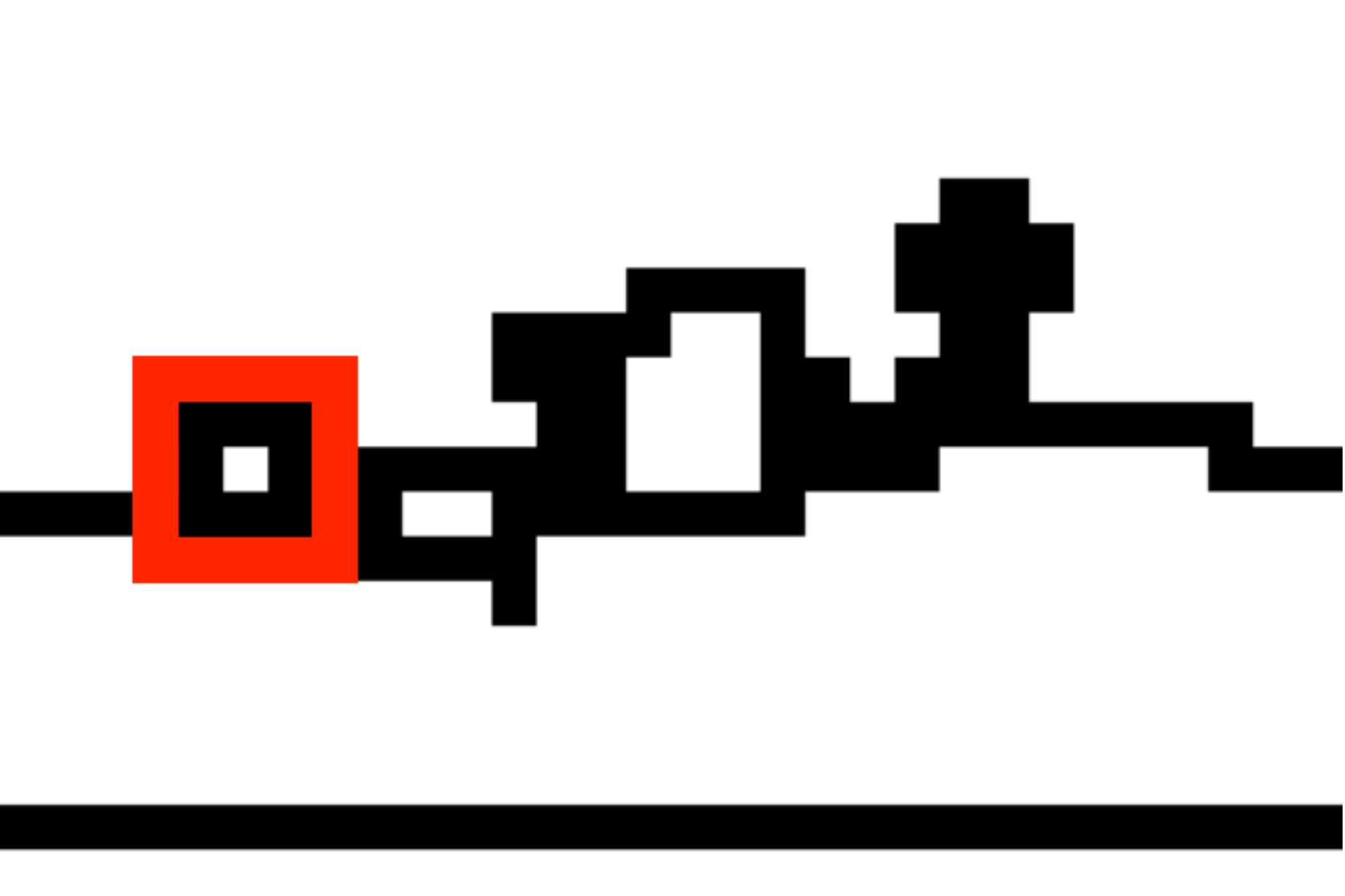}}} & &
\includegraphics[width=1.2in]{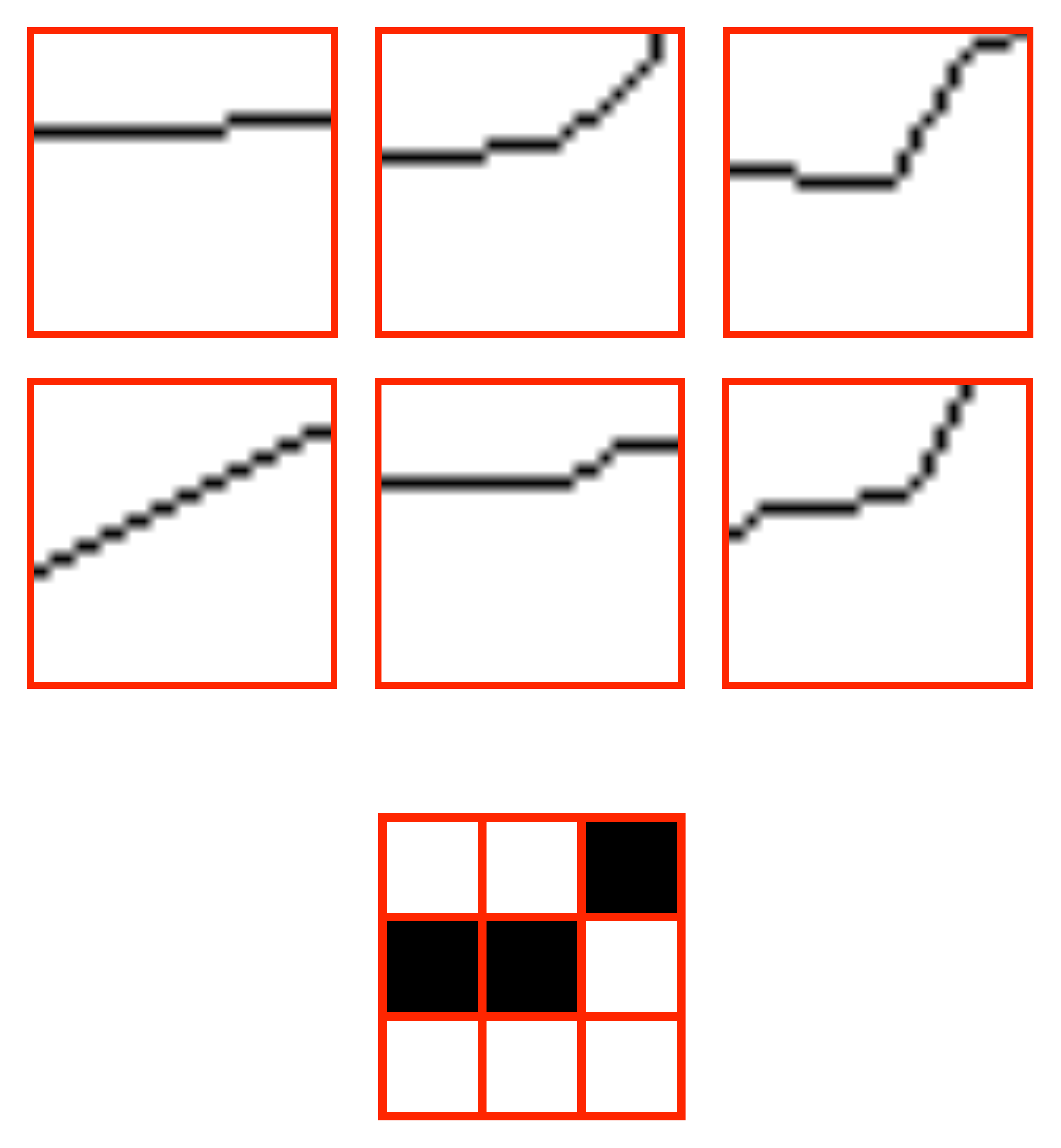} \\ \\
(a) & (b) & $\;\;\;\;\;$ & (c)
\end{tabular}
\caption{(a) Multiscale/pyramid representation of a contour map.  (b)
  Coarsest image scaled up for better visualization, with a 3x3
  pattern highlighted.  The leftmost object in the original image
  appears as a 3x3 ``circle'' pattern in the coarse image.  (c)
  Patches of contour maps (top) that coarsen to a particular 3x3
  pattern (bottom) after reducing their resolution by a factor of 8.}
\label{fig:coarsening}
\end{figure}

We define a \emph{Field of Patterns} (FoP) model using an energy
function that assigns a cost to each 3x3 pattern observed in an image
pyramid.  The cost of a pattern depends on the scale where it appears.
Figure~\ref{fig:coarsening} shows a binary image corresponding to a
contour map from the Berkeley segmentation dataset (BSD)
\cite{MFM04,AMFM11} and a pyramid representation obtained by repeated
coarsening.  The 3x3 patterns we observe after repeated coarsening
depend on large neighborhoods of the original image.  These coarse 3x3
patterns capture non-local image properties.  We train models using a
maximum-likelihood criteria.  This involves selecting pattern costs
making the expected frequency of patterns in a random sample from the
model match the average frequency of patterns in the training images.
Using the pyramid representation the model matches frequencies of
patterns at each resolution.


In practice we use MCMC methods for inference and learning.  In
Section~\ref{sec:band} we describe an MCMC sampling algorithm
that can update a very large area of an image (a horizontal or
vertical band of pixels) in a single step, by combining the
forward-backward algorithm for one-dimensional Markov models with a
Metropolis-Hastings procedure.

We evaluated our models and algorithms on two different applications.
One involves contour detection.  The other involves binary
segmentation.  These two applications require very different image
priors.  For contour detection the prior should encourage a network of
thin contours, while for binary segmentation the prior should
encourage spatially coherent masks.  In both cases we can design
effective models using maximum-likelihood estimation.  

\subsection{Related Work}

FRAME models \cite{ZWM98} and more recently Fields of Experts (FoE)
\cite{RB09} defined high-order energy models using the response of
linear filters.  FoP models are closely related.  The detection of 3x3
patterns at different resolutions corresponds to using
\emph{non-linear} filters of increasing size.  In FoP we have a fixed
set of pre-defined non-linear filters that detect common patterns at
different resolutions.  This avoids filter
learning, which leads to a non-convex optimization problem in FoE.


A restricted set of 3x3 binary patterns was considered in
\cite{DMPS95} to define priors for image restoration.  Binary patterns
were also used in \cite{SKR11} to model curvature of a binary shape.
There has been recent work on inference algorithms for CRFs defined by
binary patterns \cite{TK13} and it may be possible to develop
efficient inference algorithms for FoP models using those techniques.

The work in \cite{W02} defined a variety of multiresolution models for
images based on a quad-tree representation.  The quad-tree leads to
models that support efficient learning and inference via dynamic
programming, but such models also suffer from artifacts due to the
underlying tree-structure.  The work in \cite{EHW12} defined binary
image priors using deep Boltzmann machines.  Those models are based on
a hierarchy of hidden variables that is related to our multiscale
representation.  However in our case the multiscale representation is
a deterministic function of the image and does not involve extra
hidden variables as \cite{EHW12}.  The approach we take to define a
multiscale model is similar to \cite{FS07} where local properties of
subsampled signals where used to model curves.

One of our motivating applications involves detecting contours in
noisy images.  This problem has a long history in computer vision,
going back at least to \cite{SU88}, who used a type of Markov model
for detecting salient contours.  Related approaches include the
stochastic completion field in \cite{WJ97a,WJ97b},
spectral methods \cite{Contours03}, the curve
indicator random field \cite{AZ03}, and the recent work in \cite{AGNB10}.

\section{Fields of Patterns (FoP)}

Let ${\cal G} = [n] \times [m]$ be the grid of pixels in an $n$ by $m$
image.  Let $x = \{x(i,j) \;|\; (i,j) \in {\cal G}\}$ be a hidden
binary image and $y = \{y(i,j) \;|\; (i,j) \in {\cal G}\}$ be a set of
observations (such as a grayscale or color image).
Our goal is to estimate $x$ from $y$.

We define a CRF $p(x|y)$ using an energy function that is a
sum of two terms,
\begin{equation}
p(x|y) =  \frac{1}{Z(y)} \exp(-E(x,y)) \;\;\;\;\;
E(x,y) = E_{\fop}(x) + E_{\data}(x,y)
\label{eqn:cond}
\end{equation}


\subsection{Single-scale FoP Model}

The single-scale FoP model is one of the simplest energy models that
can capture the basic properties of contour maps or other images that
contain thin objects.
We use $x[i,j]$ to denote the binary pattern
defined by $x$ in the 3x3 window centered at pixel $(i,j)$, treating
values outside of the image as 0.
A single-scale FoP model is defined by the
local patterns in $x$,
\begin{equation}
E_{\fop}(x) = \sum_{(i,j) \in {\cal G}} V(x[i,j]).
\end{equation} 
Here $V$ is a potential function assigning costs (or energies) to
3x3 binary patterns.  The sum is over all 3x3 windows in $x$, including
overlapping windows.
Note that there are 512 possible binary patterns in
a 3x3 window.  We can make the model invariant to rotations and mirror
symmetries by tying parameters together.  The resulting model has 102
parameters (some patterns have more symmetries than others) and can be
learned from smaller datasets.  We used invariant models for all of
the experiments reported in this paper.

\subsection{Multiscale FoP Model}

To capture non-local statistics we look at local patterns in a
multiscale representation of $x$.  For a model with $K$ scales let
$\sigma(x) = x^0,\ldots,x^{K-1}$ be an image pyramid where $x^0 = x$
and $x^{k+1}$ is a coarsening of $x^k$.  Here $x^k$ is a binary image
defined over a grid ${\cal G}^k = [n/2^k] \times [m/2^k]$.  The
coarsening we use in practice is defined by a logical OR operation,
\begin{equation}
x^{k+1}(i,j) = x^k(2i,2j) \vee x^k(2i+1,2j) \vee x^k(2i,2j+1)^k \vee x^k(2i+1,2j+1)
\end{equation}
This particular coarsening maps connected objects at one scale of
resolution to connected objects at the next scale, but other
coarsenings may be appropriate in different applications.

A multiscale FoP model is defined by the local patterns in
$\sigma(x)$,
\begin{equation}
E_{\fop}(x) = \sum_{k=0}^{K-1} \sum_{(i,j) \in {\cal G}^k}
V^k(x^k[i,j]).
\end{equation}
This model is parameterized by $K$ potential functions $V^k$,
one for each
scale in the pyramid $\sigma(x)$.  In many applications we expect the
frequencies of a 3x3 pattern to be different at each scale.  The
potential functions can encourage or discourage specific
patterns to occur at specific scales.

Note that $\sigma(x)$ is a deterministic function and the pyramid
representation does not introduce new random variables.  The pyramid
simply defines a convenient way to specify potential
functions over large regions of $x$.
A single potential function in a
multiscale model can depend on a large area of $x$ due to the
coarsenings.
For large enough $K$ (proportional to $\log$ of the
image size) the Markov blanket of a pixel can be the whole image.

\begin{figure}
\centering
\begin{tabular}{ccc}
\setlength{\fboxsep}{0pt}
\fbox{\includegraphics[height=1.1in]{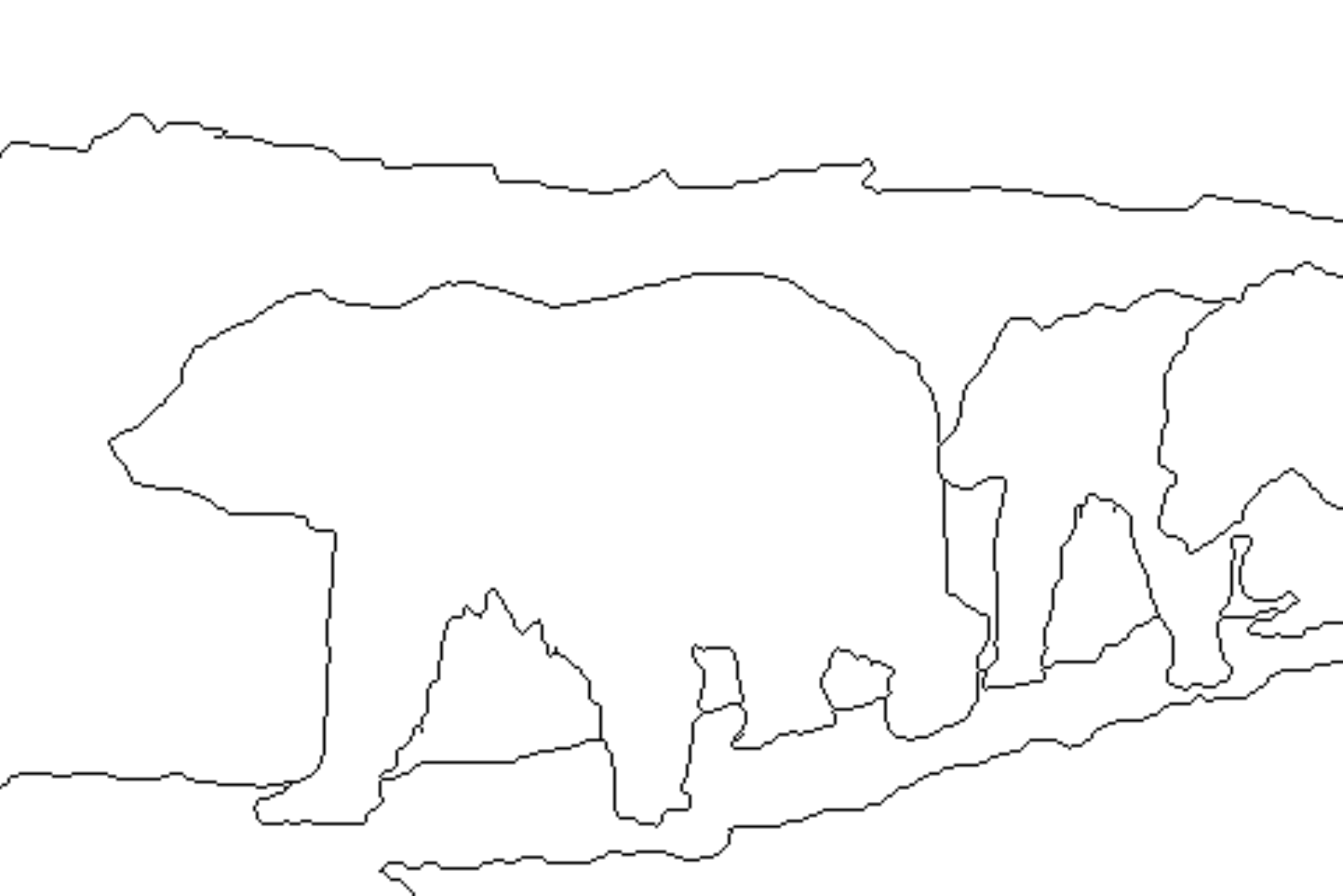}} &
\setlength{\fboxsep}{0pt}
\fbox{\includegraphics[height=1.1in]{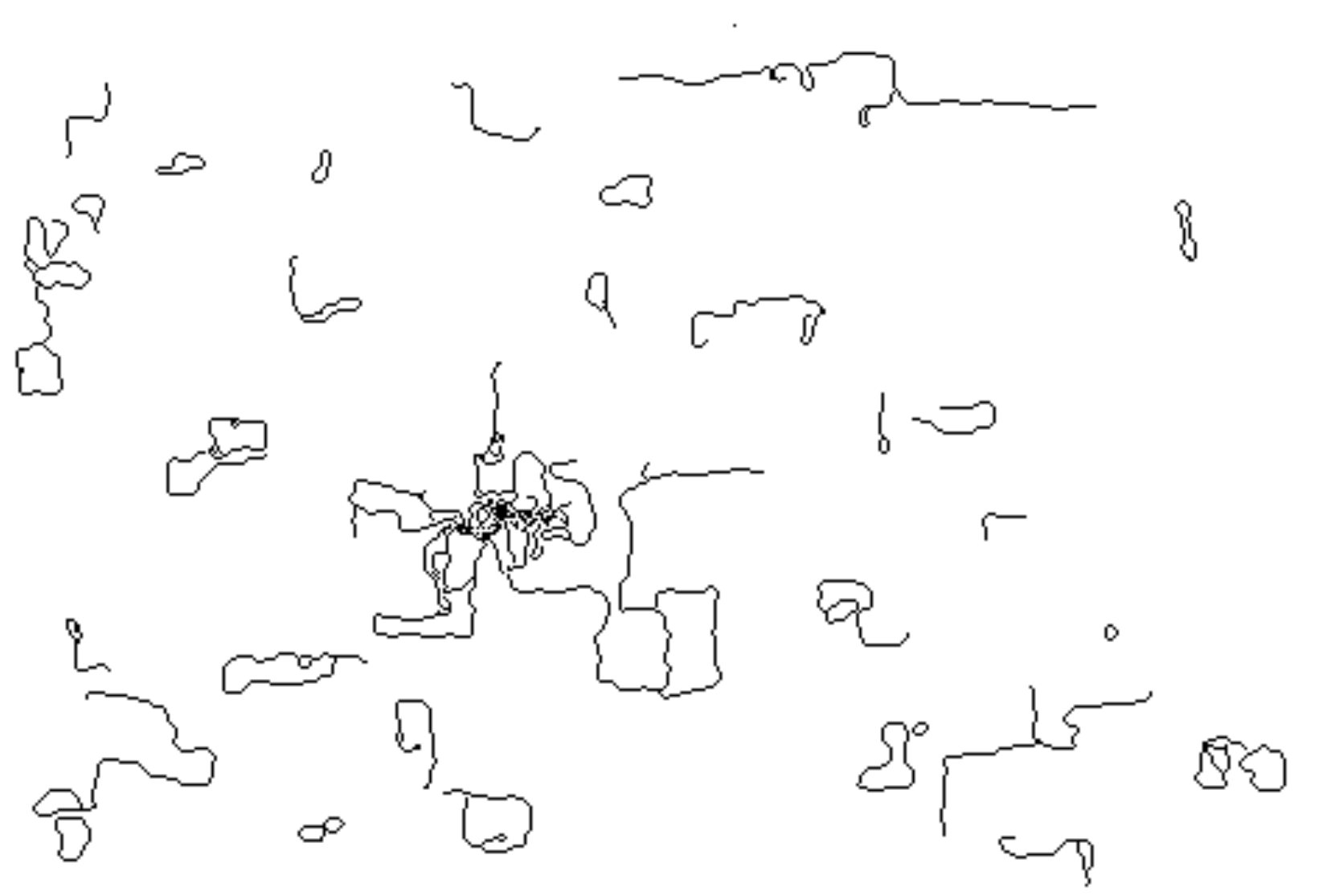}} &
\setlength{\fboxsep}{0pt}
\fbox{\includegraphics[height=1.1in]{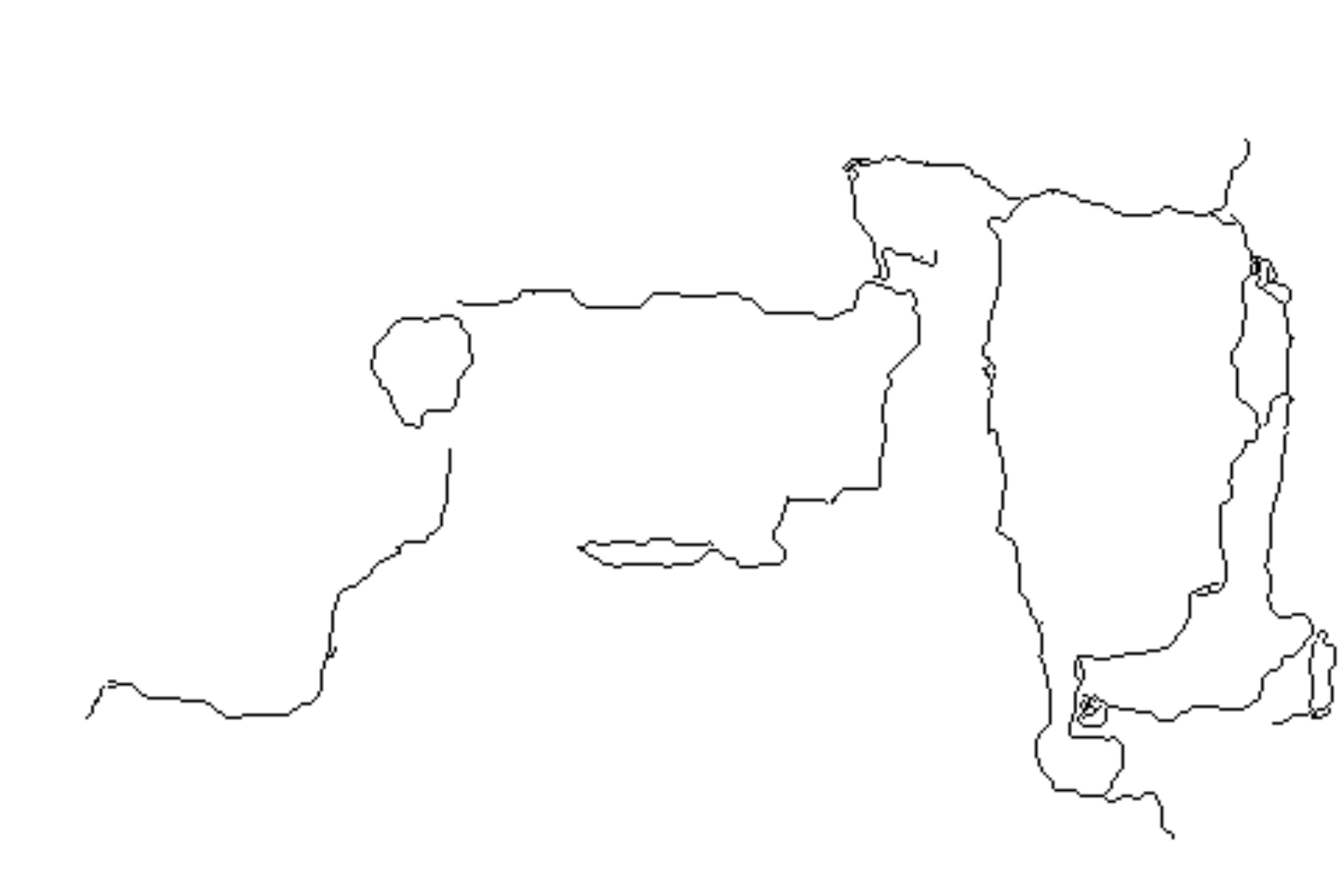}} \\
\setlength{\fboxsep}{0pt}
\fbox{\includegraphics[height=1.1in]{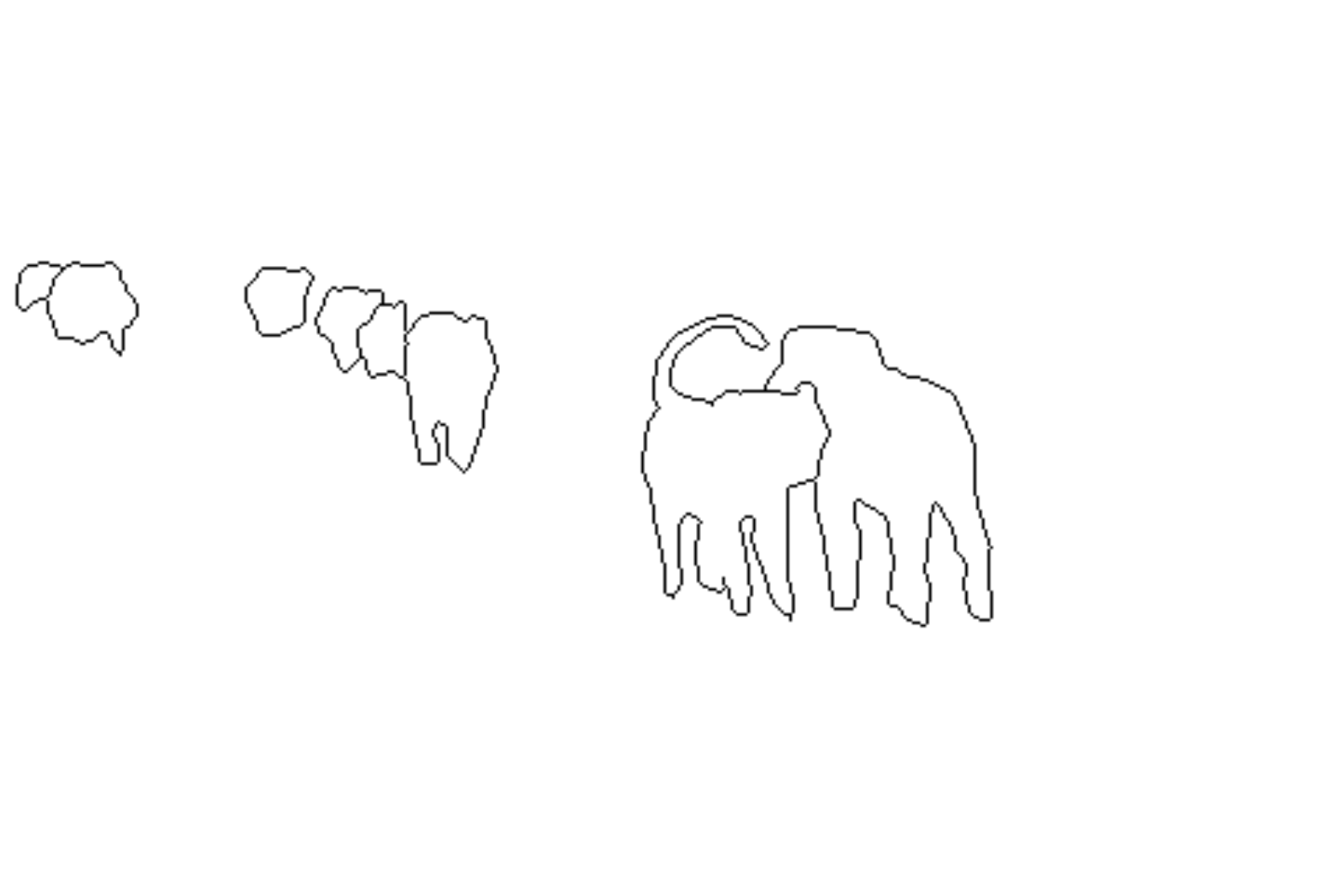}} &
\setlength{\fboxsep}{0pt}
\fbox{\includegraphics[height=1.1in]{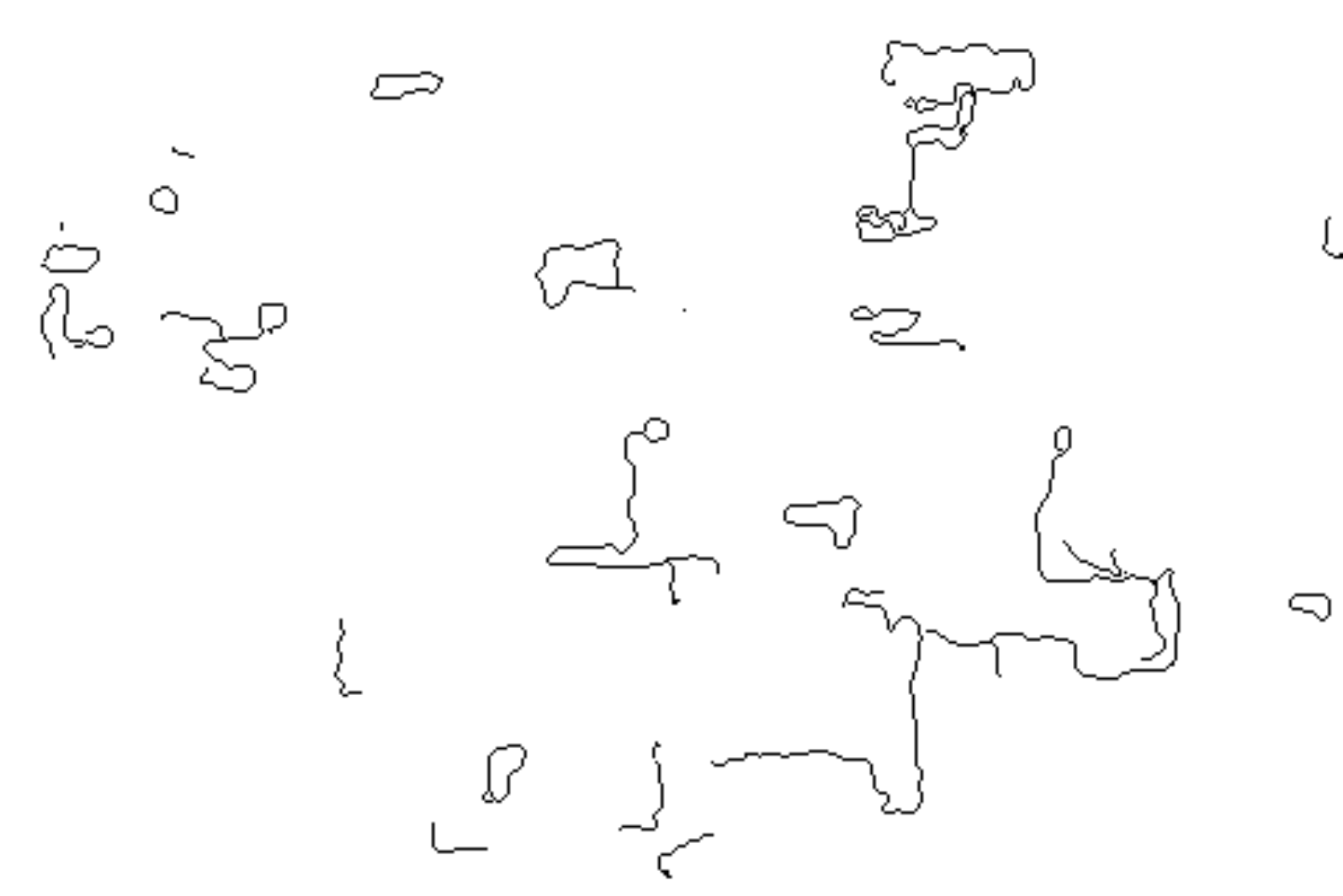}} &
\setlength{\fboxsep}{0pt}
\fbox{\includegraphics[height=1.1in]{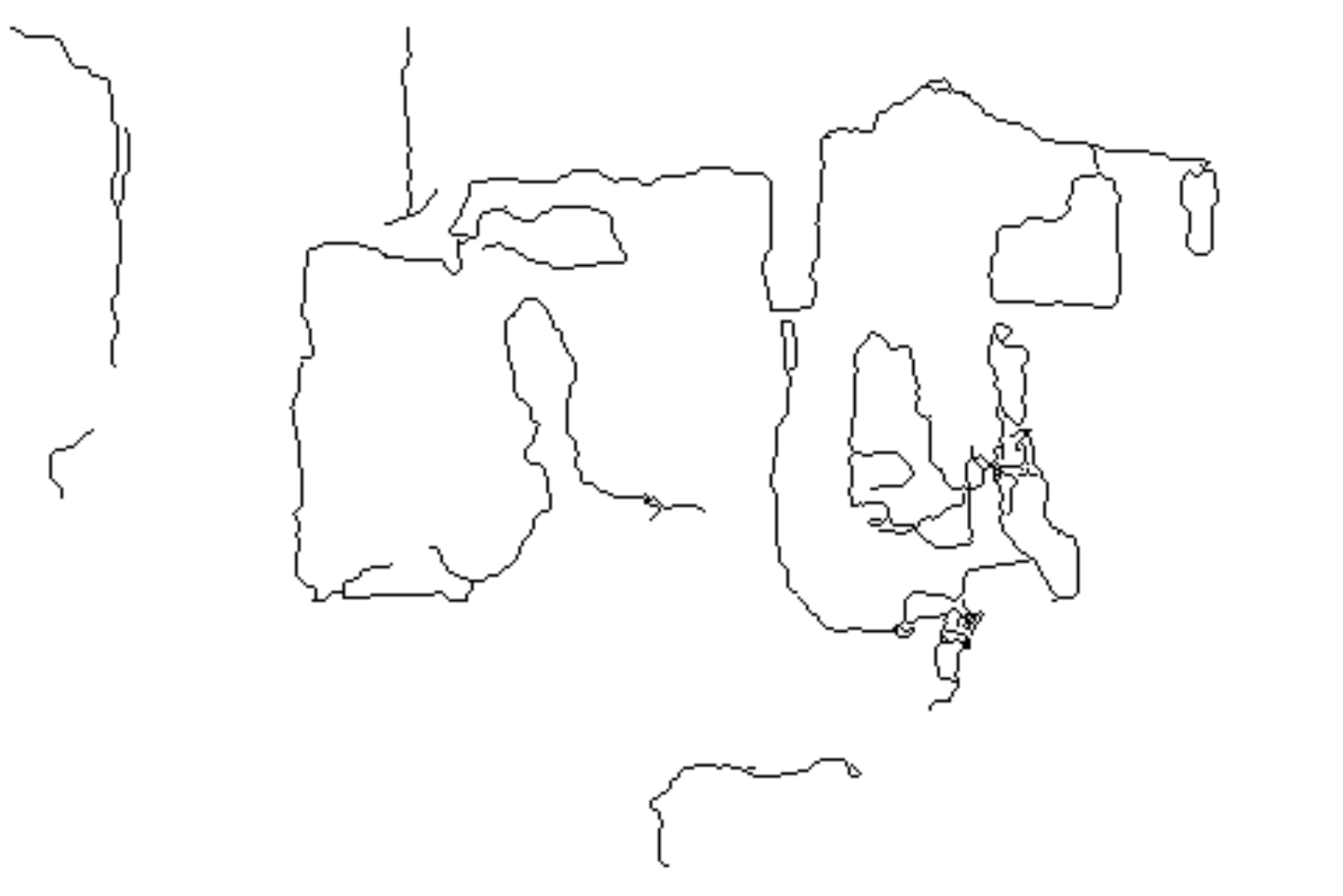}} \\
(a) & (b) & (c)
\end{tabular}
\caption{(a) Examples of training images $T$ extracted from the BSD.
  (b) Samples from a single-scale FoP prior trained on $T$.  (c)
  Samples from a multiscale FoP prior trained on $T$.  The
  multiscale model is better at capturing the lengths of
  contours and relationships between them.}
\label{fig:prior}
\end{figure}

While the experiments in Section~\ref{sec:applications} use the
conditional modeling approach specified by Equation (\ref{eqn:cond}),
we can also use $E_{\fop}$ to define priors over binary images.
Samples from these priors illustrate the information that is captured
by a FoP model, specially the added benefit of the multiscale
representation.  Figure~\ref{fig:prior} shows samples from FoP priors
trained on contour maps of natural images.

The empirical studies in \cite{RFM08} suggest that low-order Markov
models can not capture the empirical length distribution of contours
in natural images.  A multiscale FoP model can control the size
distribution of objects much better than a low-order MRF.  After
coarsening the diameter of an object goes down by a factor of
approximately two, and eventually the object is mapped to a single
pixel.  The scale at which this happens can be captured by a 3x3
pattern with an ``on'' pixel surrounded by ``off'' pixels (this
assumes there are no other objects nearby).  Since the cost of a
pattern depends on the scale at which it appears we can assign a cost
to an object that is based loosely upon its size.

\subsection{Data Model}

Let $y$ be an input image and $\sigma(y)$ be an image pyramid computed
from $y$.  Our data models are defined by sums over pixels in the two
pyramids $\sigma(x)$ and $\sigma(y)$.  In our experiments $y$ is a
graylevel image with values in $\{0,\ldots,M-1\}$.  The pyramid
$\sigma(y)$ is defined in analogy to $\sigma(x)$ except that we use a
local average for coarsening instead of the logical OR,
\begin{equation}
y^{k+1}(i,j) = \lfloor (y^k(2i,2j) + y^k(2i+1,2j) + y^k(2i,2j+1) + y^k(2i+1,2j+1))/4 \rfloor
\end{equation}
The data model is parameterized by $K$ vectors $D^0,\ldots,D^{K-1} \in \mathbb{R}^M$
\begin{equation}
E_{\data}(x,y) = \sum_{k=0}^{K-1} \sum_{(i,j) \in {\cal G}^k} x^k(i,j) D^k(y^k(i,j))
\end{equation}
Here $D^k(y^k(i,j))$ is an observation cost incurred when
$x^k(i,j) = 1$.  There is no need to include an observation cost when
$x^k(i,j) = 0$ because only energy differences 
affect the posterior $p(x|y)$.

We note that it would be interesting to consider data models that
capture complex relationships between local patterns in $\sigma(x)$
and $\sigma(y)$.  For example a local maximum in $y^k(i,j)$ might give
evidence for $x^k(i,j) = 1$, or a particular 3x3 pattern in
$x^k[i,j]$.

\subsection{Log-Linear Representation}
\label{sec:loglinear}

The energy function $E(x,y)$ of a FoP model can be expressed by a dot
product between a vector of model parameters $w$ and a feature vector
$\phi(x,y)$.  The vector $\phi(x,y)$ has one block for each scale.  In
the $k$-th block we have: (1) 512 (or 102 for invariant models)
entries counting the number of times each 3x3 pattern occurs in $x^k$;
and (2) $M$ entries counting the number of times each possible value
for $y(i,j)$ occurs where $x^k(i,j) = 1$.  
The vector $w$ specifies the cost for each pattern in each scale ($V^k$) and
the parameters of the data model ($D^k$).
We then have that $E(x,y) = w \cdot \phi(x,y)$.
This log-linear form is useful for learning the model parameters
as described in Section~\ref{sec:learning}.

\section{Inference with a Band Sampler}
\label{sec:band}

In inference we have a set of observations $y$ and want to estimate
$x$.  We use MCMC methods \cite{N93} to draw samples from $p(x|y)$ and
estimate the posterior marginal probabilities $p(x(i,j) = 1 | y)$.
Sampling is also used for learning model parameters as described in
Section~\ref{sec:learning}.

In a block Gibbs sampler we repeatedly update $x$ by picking a block
of pixels $B$ and sampling new values for $x_B$ from $p(x_B | y,
x_{\overline{B}})$.  If the blocks are selected appropriately this
defines a Markov chain with stationary distribution $p(x|y)$.

We can implement a block Gibbs sampler for a multiscale FoP model by
keeping track of the image pyramid $\sigma(x)$ as we update $x$.  To
sample from $p(x_B|y,x_{\overline{B}})$ we consider each possible
configuration for $x_B$.  We can efficiently update $\sigma(x)$ to
reflect a possible configuration for $x_B$ and evaluate the terms in
$E(x,y)$ that depend on $x_B$.  This takes $O(K|B|)$ time for each
configuration for $x_B$.  This in turn leads to an $O(K|B|2^{|B|})$
time algorithm for sampling from $p(x_B | y, x_{\bar{B}})$.  The
running time can be reduced to $O(K2^{|B|})$ using Gray codes to
iterate over configurations for $x_B$.

Here we define a \emph{band sampler} that updates \emph{all} pixels in
a horizontal or vertical band of $x$ in a single step.  Consider an
$n$ by $m$ image $x$ and let $B$ be a horizontal band of pixels with
$h$ rows.  Since $|B| = mh$ a straightforward implementation of block
sampling for $B$ is completely impractical.  However, for an
Ising model we can generate samples from $p(x_B|y,x_{\overline{B}})$
in $O(m 2^{2h})$ time using the forward-backward algorithm for Markov
models.  We simply treat each column of $B$ as a single variable with
$2^h$ possible states.  A similar idea can be used for FoP models.

Let $S$ be a state space where a state specifies a joint configuration of
binary values for the pixels in a column of $B$.  Note that $|S|=2^h$.  Let
$z_1,\ldots,z_m$ be a representation of $x_B$ in terms of the state of
each column.  For a single-scale FoP model the distribution
$p(z_1,\ldots,z_n | y, x_{\bar{B}})$ is a 2nd-order Markov model.
This allows for efficient sampling using forward weights
computed via dynamic programming.  Such an algorithm takes $O(m
2^{3h})$ time to generate a sample from $p(x_B|y,x_{\overline{B}})$,
which is efficient for moderate values of $h$.

In a multiscale FoP model the 3x3 patterns in the upper levels of
$\sigma(x)$ depend on many columns of $B$.  This means
$p(z_1,\ldots,z_n | x_{\bar{B}})$ is no longer 2nd-order.  Therefore
instead of sampling $x_B$ directly we
use a Metropolis-Hastings approach.  Let $p$ be a multiscale FoP model
we would like to sample from.  Let $q$ be a single-scale FoP model that
approximates $p$.  Let $x$ be the current state of the Markov chain
and $x'$ be a proposal generated by the single-scale band sampler for
$q$.  We accept $x'$ with probability
$\min(1,((p(x'|y)q(x|y))/(p(x|y)q(x'|y))))$.  Efficient computation of
acceptance probabilities can be done using the pyramid representations
of $x$ and $y$.  For each proposal we update $\sigma(x)$ to
$\sigma(x')$ and compute the difference in energy due to the change
under both $p$ and $q$.

One problem with the Metropolis-Hastings approach is that if proposals
are rejected very often the resulting Markov chain mixes slowly.  We
can avoid this problem by noting that most of the work required to
generate a sample from the proposal distribution involves computing
forward weights that can be re-used to generate other samples.  Each
step of our band sampler for a multiscale FoP model picks a band $B$
(horizontal or vertical) and generates many proposals for $x_B$,
accepting each one with the appropriate acceptance probability.  As
long as one of the proposals is accepted the work done in computing
forward weights is not wasted.

\section{Learning}
\label{sec:learning}

We can learn models using maximum-likelihood and stochastic gradient
descent.  This is similar to what was done in \cite{ZWM98,RB09,T08}.
But in our case we have a conditional model so we maximize the
conditional likelihood of the training examples.

Let $T =
\{(x_1,y_i),\ldots,(x_N,y_N)\}$ be a training set with $N$ examples.
We define the training objective using the negative log-likelihood of
the data plus a regularization term.  The regularization ensures no
pattern is too costly.  This helps the Markov chains used during
learning and inference to mix reasonably fast.  Let $L(x_i,y_i) =
-\log p(x_i|y_i)$.  The training objective is given by
\begin{equation}
O(w) = \frac{\lambda}{2} ||w||^2 + \sum_{i=1}^N L(x_i,y_i).
\end{equation}
This objective is convex and
\begin{equation}
\nabla O(w) = \lambda w + \sum_{i=1}^N \phi(x_i,y_i) - E_{p(x|y_i)}[\phi(x,y_i)].
\end{equation}
Here $E_{p(x|y_i)}[\phi(x,y_i)]$ is the expectation of $\phi(x,y_i)$
under the posterior $p(x|y_i)$ defined by the current model parameters
$w$.  A stochastic approximation to the gradient $\nabla O(w)$ can be
obtained by sampling $x_i'$ from $p(x|y_i)$.  Let $\eta$ be a learning
rate.  In each stochastic gradient descent step we sample $x_i'$ from
$p(x|y_i)$ and update $w$ as follows
\begin{equation}
w := w - \eta(\lambda w + \sum_{i=1}^N \phi(x_i,y_i) -
\phi(x_i',y_i)).
\end{equation} 
To sample the $x_i'$ we run $N$ Markov chains, one for each training
example, using the band sampler from Section~\ref{sec:band}.  After
each model update we advance each Markov chain for a small number of
steps using the latest model parameters to obtain new samples $x_i'$.

\section{Applications}
\label{sec:applications}

To evaluate the ability of FoP to adapt to different problems we
consider two different applications.  In
both cases we estimate hidden binary images $x$ from grayscale input
images $y$.  We used ground-truth binary images $x$ from standard
datasets and synthetic observations $y$.  For the experiments described
here we generated $y$ by sampling a value $y(i,j)$ for each pixel
independently from a
normal distribution with standard deviation $\sigma_y$ and mean
$\mu_0$ or $\mu_1$, depending on $x(i,j)$,
\begin{equation}
y(i,j) \sim {\cal N}(\mu_{x(i,j)},\sigma_y^2).
\label{eqn:obs}
\end{equation}
We have also done experiments with more complex observation
models but the results we obtained were similar to the results
described here.

\subsection{Contour Detection}

The BSD \cite{MFM04,AMFM11} contains images of natural scenes and
manual segmentations of the most salient objects in those images.  We
used one manual segmentation for each image in the BSD500.  From each
image we generated a contour map $x$ indicating the location of
boundaries between segments.  To generate the
observations $y$ we used $\mu_0=150$, $\mu_1=100$ and $\sigma_y = 40$
in Equation (\ref{eqn:obs}).  Our training and test sets each have 200
examples.  We first trained a 1-scale FoP model.  We then trained a
4-level FoP model using the 1-level model as a proposal distribution
for the band sampler (see Section~\ref{sec:band}).  Training each
model took 2 days on a 20-core machine.  During training and testing
we used the band sampler with $h=3$ rows.  Inference involves
estimating posterior probabilities for each pixel by sampling
from $p(x|y)$.  Inference on each image took 20 minutes on an 8-core
machine.

\begin{figure}
\centering
\renewcommand{\tabcolsep}{2pt}
\begin{tabular}{lccc}
\rotatebox{90}{\hspace{4mm}Contour map $x$} &
\setlength{\fboxsep}{0pt}
\fbox{\includegraphics[width=1.65in]{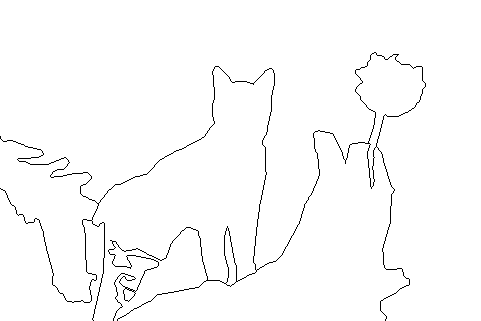}} &
\setlength{\fboxsep}{0pt}
\fbox{\includegraphics[width=1.65in]{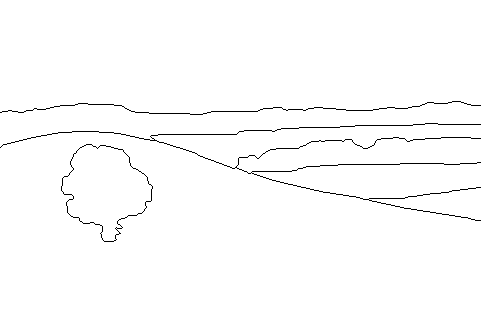}} &
\setlength{\fboxsep}{0pt}
\fbox{\includegraphics[width=1.65in]{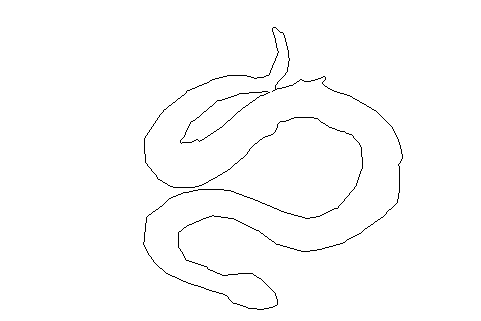}} \\
\rotatebox{90}{\hspace{5mm}Observation $y$} &
\setlength{\fboxsep}{0pt}
\fbox{\includegraphics[width=1.65in]{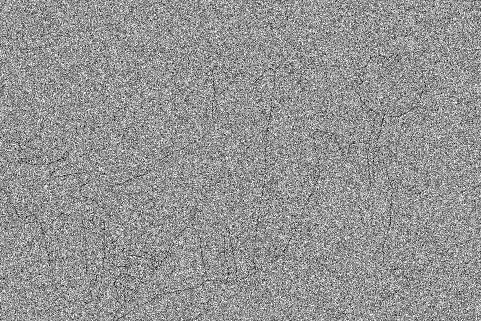}} &
\setlength{\fboxsep}{0pt}
\fbox{\includegraphics[width=1.65in]{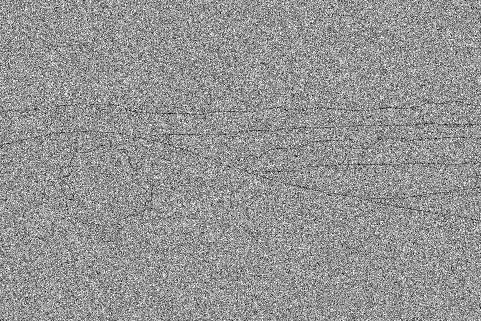}} &
\setlength{\fboxsep}{0pt}
\fbox{\includegraphics[width=1.65in]{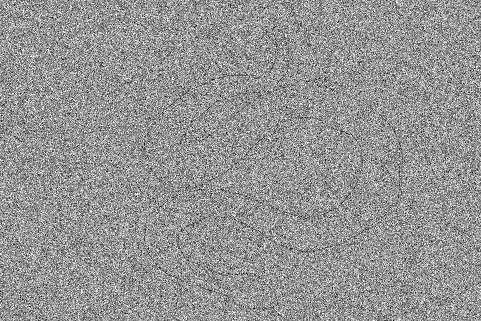}} \\
\rotatebox{90}{\hspace{3mm}Baseline $\sigma_b = 1$} &
\setlength{\fboxsep}{0pt}
\fbox{\includegraphics[width=1.65in]{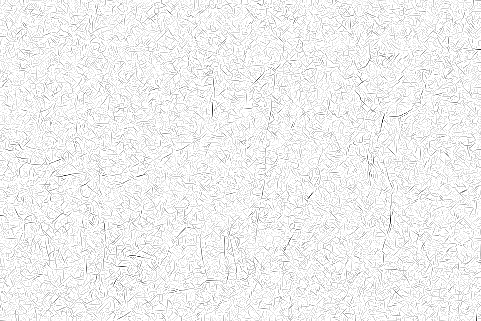}} &
\setlength{\fboxsep}{0pt}
\fbox{\includegraphics[width=1.65in]{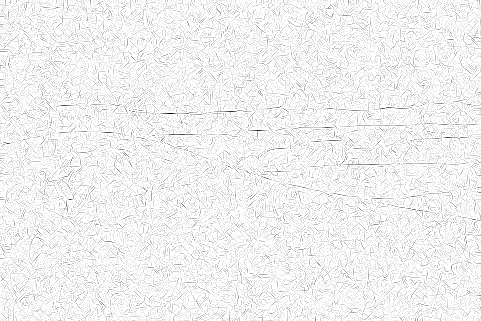}} &
\setlength{\fboxsep}{0pt}
\fbox{\includegraphics[width=1.65in]{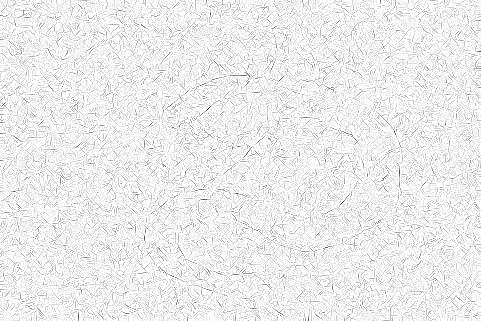}} \\
\rotatebox{90}{\hspace{3mm}Baseline $\sigma_b = 4$} &
\setlength{\fboxsep}{0pt}
\fbox{\includegraphics[width=1.65in]{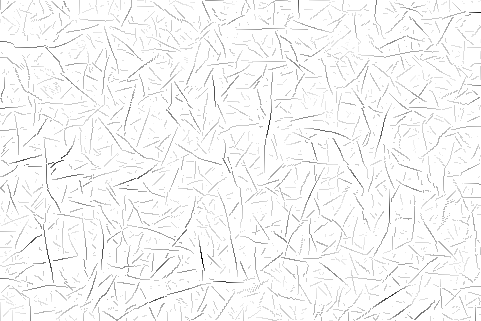}} &
\setlength{\fboxsep}{0pt}
\fbox{\includegraphics[width=1.65in]{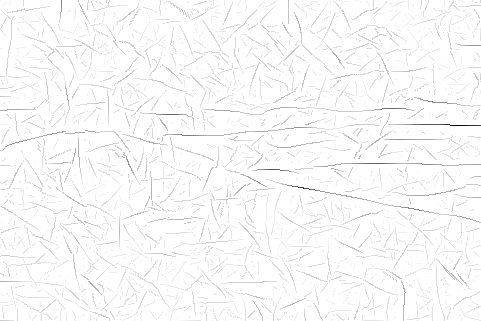}} &
\setlength{\fboxsep}{0pt}
\fbox{\includegraphics[width=1.65in]{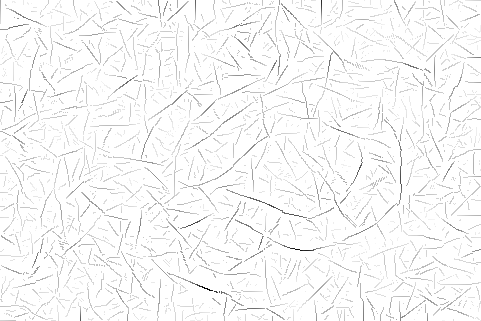}} \\
\rotatebox{90}{\hspace{10mm}FoP 1} &
\setlength{\fboxsep}{0pt}
\fbox{\includegraphics[width=1.65in]{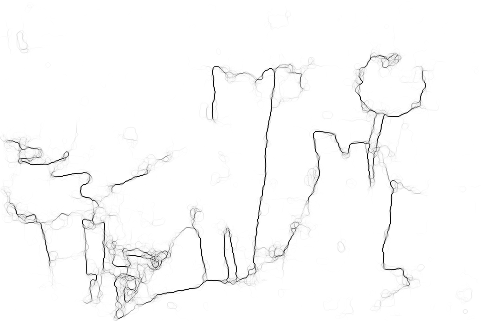}} &
\setlength{\fboxsep}{0pt}
\fbox{\includegraphics[width=1.65in]{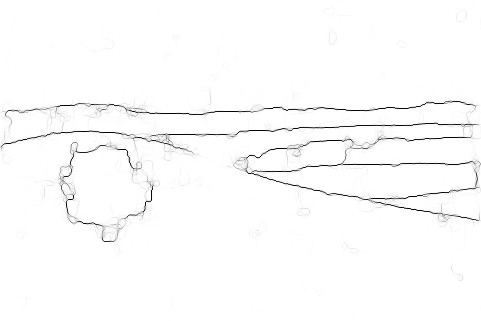}} &
\setlength{\fboxsep}{0pt}
\fbox{\includegraphics[width=1.65in]{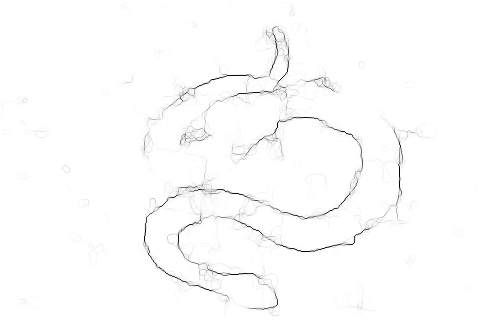}} \\
\rotatebox{90}{\hspace{10mm}FoP 4} &
\setlength{\fboxsep}{0pt}
\fbox{\includegraphics[width=1.65in]{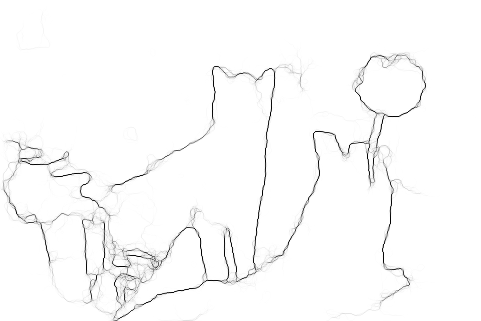}} &
\setlength{\fboxsep}{0pt}
\fbox{\includegraphics[width=1.65in]{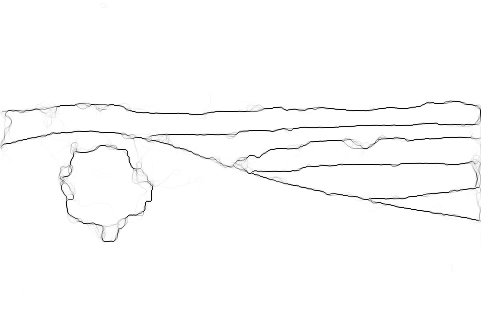}} &
\setlength{\fboxsep}{0pt}
\fbox{\includegraphics[width=1.65in]{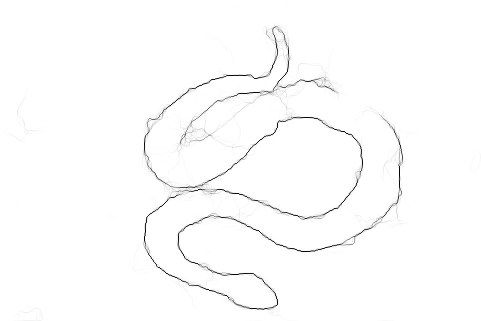}} 
\end{tabular}
\caption{Contour detection results.  Top-to-bottom: Hidden contour map
  $x$, input image $y$, output of oriented filter baseline with
  $\sigma_b = 1$ and $\sigma_b = 4$, output of 1-level and 4-level FoP
  model.}
\label{fig:BSD}
\end{figure}

For comparison we implemented a baseline technique using linear
filters.  Following \cite{LM98} we used the second derivative of an
elongated Gaussian filter together with its Hilbert transform.  The
filters had an elongation factor of 4 and we experimented with
different values for the base standard deviation $\sigma_b$ of the
Gaussian.  The sum of squared responses of both filters defines an
oriented energy map.  We evaluated the filters at 16 orientations and
took the maximum response at each pixel.  We performed non-maximum
suppression along the dominant orientations to obtain a thin
contour map.

Figure~\ref{fig:BSD} illustrates our results on 3 examples from the
test set.  Results on more examples are available in the supplemental
material.  For the FoP models we show the posterior probabilities for
each pixel $p(x(i,j) = 1|y)$.  The darker pixels have higher posterior
probability.  The FoP models do a good job suppressing
noise and localizing the contours.  The multiscale FoP model in
particular gives fairly clean results despite the highly noisy inputs.
The baseline results at lower $\sigma_b$ values suffer from
significant noise, detecting many spurious edges.  The baseline at
higher $\sigma_b$ values suppresses noise at the expense of having
poor localization and missing high-curvature boundaries.

\begin{figure}
\centering
\begin{tabular}{cc}
\includegraphics[width=2.6in]{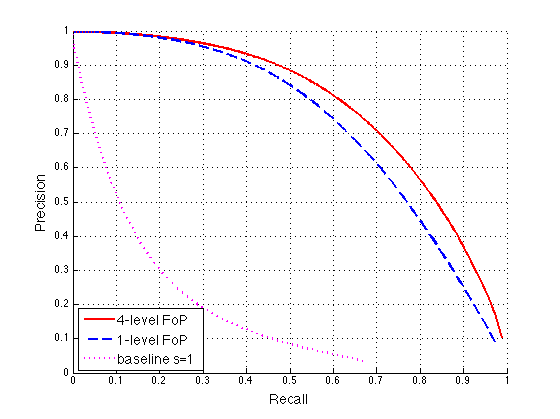} &
\includegraphics[width=2.6in]{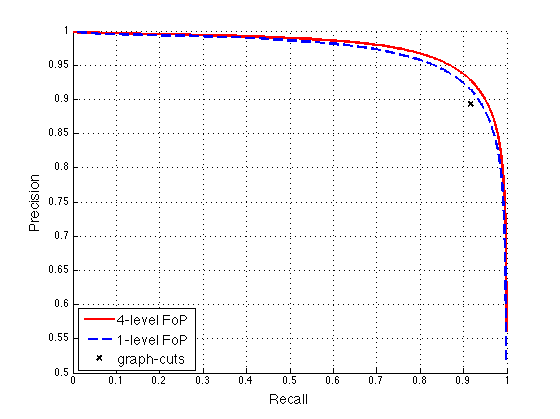} \\
(a) Contour detection & (b) Binary segmentation
\end{tabular}
\caption{(a) Precision-recall curves for the contour detection
  experiment.  (b) Precision-recall curves for the segmentation
  experiment (the graph-cuts baseline yields a single precision-recall
  point).}
\label{fig:PR}
\end{figure}

For a quantitative evaluation we compute precision-recall curves for
the different models by thresholding the estimated contour maps at
different values.  Figure~\ref{fig:PR} shows the precision-recall
curves.  The average precision (AP) was found by calculating the area
under the precision-recall curves.  The 1-level FoP model AP was 0.73.
The 4-level FoP model AP was 0.78.  The best baseline AP was 0.18
obtained with $\sigma_b = 1$.  We have also done experiments using
lower observation noise levels $\sigma_y$.  With low observation noise
the 1-level and 4-level FoP results become similar and baseline
results improve significantly approaching the FoP results.

\subsection{Binary Segmentation}

For this experiment we obtained binary images from the Swedish Leaf
Dataset \cite{S01}.  We focused on the class of Rowan leaves because
they have complex shapes.  Each image defines a segmentation mask $x$.
To generate the observations $y$ we used $\mu_0 = 150$, $\mu_1 = 100$
and $\sigma_y = 100$ in Equation (\ref{eqn:obs}).  We used a higher
$\sigma_y$ compared to the previous experiment because the 2D nature
of masks makes it possible to recover them under higher noise.  We
used $50$ examples for training and $25$ examples for testing.  We
trained FoP models with the same procedure and parameters used for the
contour detection experiment.  For a baseline, we used graph-cuts
\cite{BVZ01,BK04} to perform MAP inference with an Ising model.  We
set the data term using our knowledge of the observation model and
picked the pairwise discontinuity cost minimizing the per-pixel error
rate in the test set.

\begin{figure*}
\centering
\renewcommand{\tabcolsep}{2pt}
\begin{tabular}{ccccc}
\small Mask $x$ &
\small Observation $y$ &
\small Graph-cuts &
\small FoP 1 &
\small FoP 4 \\
\setlength{\fboxsep}{0pt}
\fbox{\includegraphics[width=1in]{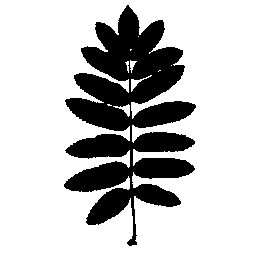}} &
\setlength{\fboxsep}{0pt}
\fbox{\includegraphics[width=1in]{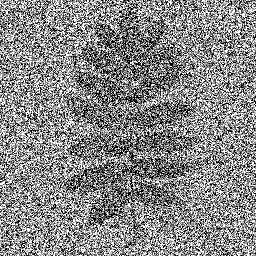}} &
\setlength{\fboxsep}{0pt}
\fbox{\includegraphics[width=1in]{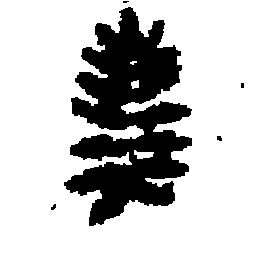}} &
\setlength{\fboxsep}{0pt}
\fbox{\includegraphics[width=1in]{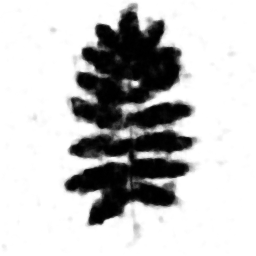}} &
\setlength{\fboxsep}{0pt}
\fbox{\includegraphics[width=1in]{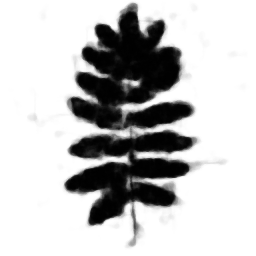}} \\
\setlength{\fboxsep}{0pt}
\fbox{\includegraphics[width=1in]{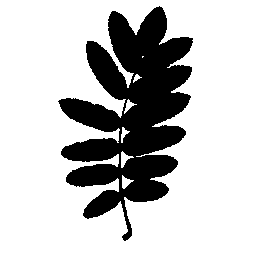}} &
\setlength{\fboxsep}{0pt}
\fbox{\includegraphics[width=1in]{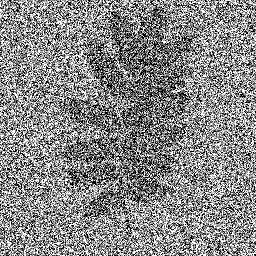}} &
\setlength{\fboxsep}{0pt}
\fbox{\includegraphics[width=1in]{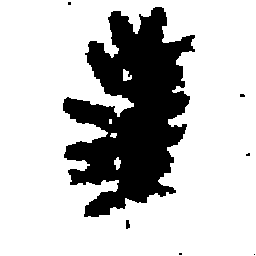}} &
\setlength{\fboxsep}{0pt}
\fbox{\includegraphics[width=1in]{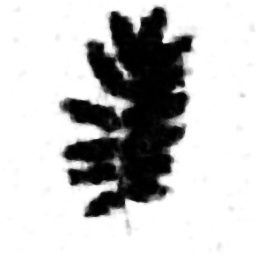}} &
\setlength{\fboxsep}{0pt}
\fbox{\includegraphics[width=1in]{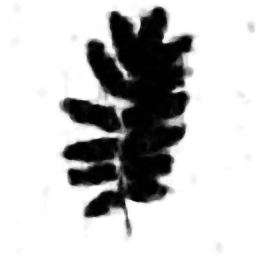}}
\end{tabular}
\caption{Binary segmentation examples.  The 4-level FoP model does a
  better job recovering pixels near the object boundary and the stem
  of the leaves.}
\label{fig:fernPlate}
\end{figure*}

Figure~\ref{fig:fernPlate} illustrates the results of the different
methods.  Results on other images are available in the supplemental
material.  The precision-recall curves are in Figure~\ref{fig:PR}.
Graph-cuts yields a precision-recall point, with precision 0.893 and
recall 0.916.  The 1-level FoP model has a higher precision of 0.915
at the same recall.  The 4-level FoP model raises the precision to
0.929 at the same recall.  The differences in precision are small
because they are due to pixels near the object boundary but those are
the hardest pixels to get right.  In practice the 4-level FoP model
recovers more detail when compared to graph-cuts.  This
can be seen by visual inspection of the leaf stem and boundaries.

\section{Conclusion}

We described a general framework for defining high-order image models.
The idea involves modeling local properties in a multiscale
representation of an image.  This leads to a natural low-dimensional
parameterization for high-order models that exploits standard pyramid
representations of images.  Our experiments demonstrate the approach
yields good results on two applications that require very different
image priors, illustrating the broad applicability of our models.  An
interesting direction for future work is to consider FoP models for
non-binary images.

\section*{Acknowledgements}

We would like to thank Alexandra Shapiro for helpful discussions and
initial experiments related to this project.  This material is based
upon work supported by the National Science Foundation under Grant
No. 1161282.

{\small
\bibliographystyle{plain}
\bibliography{fop}
}

\end{document}